\documentclass[lettersize,journal]{IEEEtran}
\usepackage{amsmath,amsfonts}
\usepackage{algorithmic}
\usepackage{algorithm}
\usepackage{array}
\usepackage[caption=false,font=normalsize,labelfont=sf,textfont=sf]{subfig}
\usepackage{textcomp}
\usepackage{stfloats}
\usepackage{url}
\usepackage{verbatim}
\usepackage{graphicx}
\usepackage{cite}
\usepackage{xcolor}
\usepackage{hyperref}
\begin{document}

\title{Decentralized Navigation of a Cable-Towed Load using Quadrupedal Robot Team via MARL}

\author{
Wen-Tse Chen$^*$, Minh Nguyen$^*$, Zhongyu Li$^*$, Guo Ning Sue, and Koushil Sreenath
\thanks{W. T. C., G. N. S. are with the Robotics Institute, Carnegie Mellon University, Pittsburgh, USA. (e-mail: wentsec,gsue@andrew.cmu.edu).}
\thanks{M. N., Z. L., and K. S. are with the  Department of Mechanical Engineering, University of California, Berkeley, Berkeley, USA (e-mail:  minh02, zhongyu\_li, koushils@berkeley.edu).}
}



\maketitle

\begin{abstract}
This work addresses the challenge of enabling a team of quadrupedal robots to collaboratively tow a cable-connected load through cluttered and unstructured environments while avoiding obstacles. 
Leveraging cables allows the multi-robot system to navigate narrow spaces by maintaining slack when necessary. 
However, this introduces hybrid physical interactions due to alternating taut and slack states, with computational complexity that scales exponentially as the number of agents increases.
To tackle these challenges, we developed a scalable and decentralized system capable of dynamically coordinating a variable number of quadrupedal robots while managing the hybrid physical interactions inherent in the load-towing task. 
At the core of this system is a novel multi-agent reinforcement learning (MARL)-based planner, designed for decentralized coordination. The MARL-based planner is trained using a centralized training with decentralized execution (CTDE) framework, enabling each robot to make decisions autonomously using only local (ego) observations.
To accelerate learning and ensure effective collaboration across varying team sizes, we introduce a tailored training curriculum for MARL. 
Experimental results highlight the flexibility and scalability of the framework, demonstrating successful deployment with one to four robots in real-world scenarios and up to twelve robots in simulation. 
The decentralized planner maintains consistent inference times, regardless of the team size.
Additionally, the proposed system demonstrates robustness to environment perturbations and adaptability to varying load weights. 
This work represents a step forward in achieving flexible and efficient multi-legged robotic collaboration in complex and real-world environments.
\end{abstract}


\section{Introduction}

\begin{figure}[!t]
\centering
\includegraphics[width=0.99\linewidth]{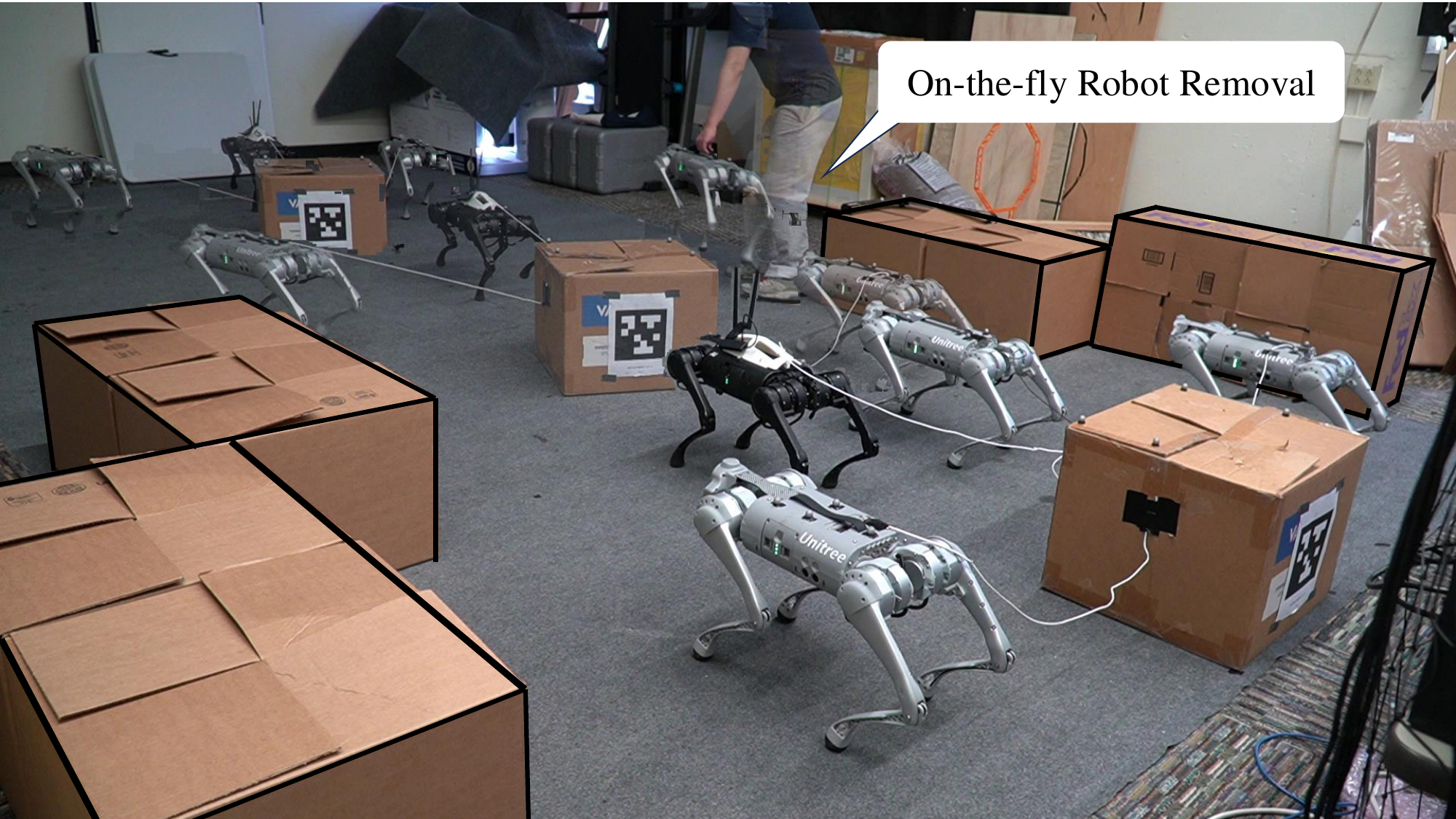}
\caption{Overlaid snapshots of four quadrupedal robots navigating a cable-towed load through a narrow passage using the proposed decentralized reinforcement-learning-based planner in this work. Later frames are made more transparent to highlight progression. Completing this task requires real-time coordination among teammates to adjust cable tension and avoid obstacles while navigating to the goal. 
The same MARL-based decentralized policy is used to flexibly control different individual robots in different team sizes. This experiment demonstrates the adaptability of the proposed decentralized policy as one robot is removed partway through the task. More experiments are shown in the \href{https://youtu.be/GkGldcfQi9k}{video}.}
\label{fig_1}
\end{figure}

\IEEEPARstart{I}{n} many real-world robotics tasks, the capabilities of a single robot could be insufficient to achieve complex tasks. 
For example, tasks such as the movement and assembly of large-scale and heavy objects often require the collaborative efforts of multiple robots~\cite{drew2021multi}. 
By working together, a robot team can accomplish tasks that go beyond the capacity of an individual robot.
Legged robots, such as quadrupedal robots, being dynamically stable systems, are inherently more robust to external loads compared to mobile robots, making them particularly suitable for load-towing tasks. 
When navigating heavy loads in cluttered and unstructured environments, cable-towed systems offer significant advantages~\cite{xiao2021robotic}. 
Cables provide flexibility in team configurations, enabling robots to maneuver through narrow spaces by dynamically switching between taut and slack states as required.
This motivates us to tackle the problem of enabling a team of quadrupedal robots to collaboratively tow a cable-connected load, navigate through complex environments, and reach a destination while avoiding obstacles, as shown in Fig.~\ref{fig_1}.

However, this problem is challenging. 
First, multiple legged robots are physically connected to a load, coupling their movements with the load and creating a complex dynamic system involving the entire robot-towing-load system. 
Additionally, the cables introduce hybrid dynamics as they can switch between taut and slack modes, further complicating the system. 
The complexity of multi-robot systems also grows exponentially with the number of robots.
This complexity makes decentralized planning and execution critical for achieving efficient real-time operation that scales with the number of robots in the team. 
The navigation task also complicates the challenge, requiring collision-free planning over a long horizon while considering obstacles, teammates, and the load.
Additionally, the robot team needs to adapt to varying load weights and changing team sizes, necessitating a planning strategy that is both flexible and adaptive.
While some efforts have been made to address multi-robot collaboration, prior work often focuses on navigation tasks without accounting for hybrid interactions among robots~\cite{9316033} or restricts the scope to teams with a small, fixed number of robots~\cite{10281391}. 
Recent work has explored centralized model predictive control for similar robot-towing-load systems~\cite{9830869}, but such an approach suffers from exponential growth in computational complexity, making it unscalable for larger teams, \emph{e.g.}, the computing time for each replanning cycle for a 12-agent team requires over 20 seconds.

In this work, we propose a decentralized MARL planner based on centralized training with decentralized execution (CTDE) framework. 
It is designed to emphasize (1) scalability, enabling the planner to handle an increasing number of agents, and (2) flexibility, allowing it to adapt to changes in team size during deployment. 
The core of this approach is \emph{a unified planning policy} that operates effectively for individual agents, regardless of team size, by conditioning strategy decisions on each robot's local (ego) observations. 
Trained across various team sizes, the decentralized policy generalizes well, ensuring both scalability and flexibility. 
As illustrated in Fig.~\ref{fig_1}, all agents utilize the same decentralized policy, achieving consistent performance and inference time irrespective of team size. This design enables dynamic reconfiguration, such as adapting seamlessly when an agent is removed mid-task, while maintaining continued progress toward the goal.

The core contribution of this work lies in developing the unified planning policy that is both scalable and flexible to varying team sizes. 
We propose a MARL-based framework within the CTDE paradigm, which trains decentralized planners to generate commands for the locomotion controllers of individual quadrupedal robots. 
To address the challenge of training for varying team sizes, we introduce a novel multi-stage training strategy for MARL, whose effectiveness is thoroughly evaluated. 
The resulting decentralized planner demonstrates scalability by successfully controlling teams of one to four robots in real-world tasks and up to twelve robots in simulation. 
In real-world experiments, our planner achieves an inference time per replanning step that is over 33 times faster than the previous centralized planning method~\cite{9830869} (0.039 seconds versus 1.306 seconds) for the same collaborative load-towing tasks and has the consistent time for more number of agents. This underscores the advantages of the proposed decentralized planning. 
Furthermore, using the proposed MARL framework, we realize several novel applications for collaborative load-towing tasks, including navigating narrow gaps and demonstrating robustness to external perturbations, changes in team size, and varying loads. 
This work presents a framework and algorithm that marks a step toward flexible, scalable, and robust collaborative systems for multi-legged robot teams in real-world environments.

\section{Related Works}
In this section, we discuss the two primary approaches to multi-robot collaboration: (1) model-based optimization and (2) multi-agent reinforcement learning (MARL). We review prior work with an emphasis on applications to real-world multi-robot tasks. Additionally, we discuss the design of MARL algorithms to provide a basis for comparison with the algorithm developed in this work.

\subsection{Model-Based Optimization in Multi-Robot Collaboration}

Model-based optimization can be divided into two categories, centralized Model Predictive Control (MPC) and decentralized MPC. Centralized approaches have been widely used to solve multi-robot collaboration tasks, especially in area coverage, exploration~\cite{miller2020mine}, and formation control~\cite{1339386,nfaileh2022formation}. Although effective, the centralized method faces the problem of scaling up.
As the number of robots increases, the system's complexity grows exponentially, posing significant challenges for centralized methods. To address this, prior work has explored techniques such as simplified dynamics models~\cite{de2023centralized} or parallelized optimization~\cite{9830869} to improve planning and control efficiency. However, centralized approaches remain limited in scalability. For example, the parallelized optimization in \cite{9830869} achieves a replanning time of less than three seconds. It supports real-time planning for up to six robots, but it fails to scale beyond that. In contrast, our approach employs a fully decentralized planner with inference times that remain invariant regardless of the number of robots.

On the other hand, decentralized model predictive control has also been explored to achieve scalability for collaborative tasks. Turrisi et al.~\cite{turrisi2024pacc} proposed a decentralized MPC with passive arms for the carrying and navigation of collaborative payloads. Zhang et al.~\cite{10610348} introduced a distributed MPC framework for a quadrotor-quadruped system to manipulate cable-towed payloads. Furthermore, hierarchical MPC frameworks, such as those in~\cite{10156031} and~\cite{10341785}, combine centralized and decentralized approaches. Centralized MPC controls the payload and decentralized MPC controls team agents, enabling quadrotors to collaboratively manipulate cable-suspended payloads in cluttered space.
Although decentralized MPC offers interpretability and flexibility by adding or removing constraints in an optimization format, it often relies on a centralized MPC for the payload, which typically assumes a fixed number of agents. 
This limits its flexibility in scenarios where agents may need to be added or removed mid-task. 
Moreover, a decentralized MPC requires explicit models for both the load and the agents, making it less adaptable to dynamic changes in the load. In contrast, our decentralized RL policy is inherently adaptable to changes in both team composition and load conditions in real-world tasks, offering a more robust and flexible solution.

\subsection{MARL in Multi-Robot Collaboration}
MARL has emerged as a promising approach for multi-robot collaboration in dynamic environments~\cite{han2024lifelike}. Recent advances in MARL for multi-robot systems have targeted applications such as multi-robot navigation~\cite{9316033, 10611322, 9197209}, collaborative manipulation~\cite{lee2019learning, mandi2024roco}, cooperative transport~\cite{9551481, 9201368}, and search-and-rescue missions~\cite{niroui2019deep, queralta2020collaborative}. These MARL-based approaches are valued for their flexibility and adaptability in dynamic settings. However, many studies focus mainly on tasks such as collision avoidance~\cite{zhao2021reinforcement}, formation control~\cite{kaushik2021learning}, or path planning~\cite{bayerlein2021multi}.
These approaches often neglect physical interactions among robots, as incorporating such interactions can complicate the optimization and make it computationally expensive for realtimeness.
Efforts to address physical interactions among robots ~\cite{9427049, 10281391, de2023centralized} typically assume rigid connections between agents, which simplifies the planning problem by avoiding the complexity of mode switching (\emph{e.g.}, taut or slack states in cables). 
While such assumptions make real-time computation feasible, they sacrifice the ability to navigate through narrow spaces: a capability enabled in this work by leveraging the flexibility of cable-towing systems that can dynamically switch between taut and slack modes.

This practice of simplifying physical interactions is especially widely used in the field of legged robots. For example, prior studies have explored collaboration in scenarios such as pushing objects without physical connections~\cite{nachum2019multi, xiong2024mqe, feng2024learning}. 
However, these works are typically limited to two-agent setups.
Our approach scales beyond this, handling one to four robots in real-world scenarios and more in simulation. 
Another notable work by Pandit et al.~\cite{pandit2024learning} involves one to three bipedal robots carrying a load using a rigid carrier in a decentralized MARL framework. 
However, their focus lies on locomotion control, where the decentralized policy realizes stable gaits while following given velocity commands for the load under varying load conditions. 
The robot is unaware of the state of its teammate and the environment.
This is fundamentally different from our focus on motion planning and control, which requires addressing long-horizon planning for collision avoidance and collaborative navigation, in addition to handling physical interactions for towing a load.

\subsection{MARL Algorithms}
In the context of algorithms used in MARL, CTDE is a widely adopted approach~\cite{lowe2017multi, rashid2020monotonic, foerster2018counterfactual}. While previous CTDE studies primarily focused on controlling multi-robot systems with a fixed team size~\cite{lowe2017multi, rashid2020monotonic, foerster2018counterfactual}, our work explores how the same decentralized planner can be extended to effectively control an arbitrary number of robots.
MAPPO~\cite{yu2022surprising} effectively combines CTDE with PPO~\cite{schulman2017proximal}. Its ablation studies examine the influence of various global state inputs on the value function, revealing that the inclusion of global information improves sample efficiency in game benchmarks~\cite{samvelyan2019starcraft}. In our work, we extend this finding to robotics tasks with a multi-modal observation space. Through an ablation study, we demonstrate that incorporating privileged state information not only improves value function estimation but also enhances overall performance.

Much previous work investigates the use of a decentralized planner to control teams with varying numbers of agents. For example, Mordtach et al.~\cite{mordatch2018emergence} demonstrate that decentralized planners trained using CTDE algorithms can zero-shot generalize to unseen environments with different team sizes. However, such studies are often restricted to simple multi-agent navigation tasks~\cite{lowe2017multi}, leaving their applicability to scenarios involving complex inter-robot hybrid interactions unexplored. In our work, we test these methods in a more complex robotic task and find that CTDE agents are no longer able to zero-shot generalize to different team sizes. We propose a multi-stage training pipeline to solve the problem.

Other approaches address the challenge of controlling the varying team sizes by designing a global state with invariant length, irrespective of the number of robots in the system. These methods typically train with a single centralized critic and rely on techniques such as self-attention mechanisms~\cite{hsu2021scalable} or structured policy models with predefined task priorities and global communication~\cite{shibata2023learning}.
In contrast, our approach trains multiple critic networks, each tailored to specific team sizes, with the capacity of the critic network and the size of the global state scaling proportionally to the number of robots. By integrating CTDE closely with a multi-stage training pipeline, this method enables the actor network to leverage curriculum learning.

\section{Preliminary}

In this section, we first introduce the notation used for MARL, followed by the development of the simplified 2D dynamic model for the multi-robot system with a cable-towed load.

\subsection{Multi-agent Reinforcement Learning}


We aim to learn a policy \(\pi\) capable of solving a set of \(N\) tasks, denoted as \(\{\mathcal{M}_1, \dots, \mathcal{M}_N\}\). Each task $\mathcal{M}_n$ is a Decentralized Partially Observable Markov Decision Process (Dec-POMDP)~\cite{oliehoek2016concise}, represented by the tuple $\mathcal{M}_n \mathrel{\mathop:}= (\mathcal{A}_n,S_n,U,T_n,r_n,O,G_n,\gamma_n)$, where $S_n$, $U$, $O_n$ and $\gamma_n$ are the global state space, combined action space, partially observable observation space and discount factor, respectively. 
$\mathcal{A}_n = \{1,\dots,n\}$ is the set of $n$ agents.
At each time step $t$, each agent $i\in \mathcal{A}_n$ chooses an individual action $u_i$ to form a combined action $\mathbf{u}\in \mathbf{U}$. 
The immediate reward function $r(s,\mathbf{u})$ is the received reward when taking combined action $\mathbf{u}$ at global state $s\in S_n$ and it is shared by all the agents.
$T_n(s,\mathbf{u},s'):S_n\times\mathbf{U}\times S_n\rightarrow [0,1]$ is the state-transition function, which defines the probability of the succeeding global state $s'$ after taking combined action $\mathbf{u}$ at global state $s$. In a Dec-POMDP, each agent can only have access to partially observable observations $o_i\in O$ according to the observation function $G_n(s,i): S_n\times\mathcal{A}_n\rightarrow O$. The combined action space $U$ and the partially observable observation space $O$ are shared between tasks.

We follow the CTDE pipeline. During decentralized execution, each agent has access only to its own local partial observation and does not know which specific task it is addressing.
That is, each agent uses a policy $\pi_i(u_i|o_i)$ to produce its action $u_i$ from its local observation $o_i$, where $\pi_i(\cdot|o_i)$  represents a probability distribution over actions.
During centralized training, agents have access to the global state and tasks are sampled from a distribution $p(\mathcal{M})$.
The objective of MARL is to learn a combined policy $\pi(\mathbf{u} | o_{1:n}) = \prod_{i=1}^{n} \pi_{i}(u_i | o_i)$ to maximize the discounted accumulated reward $\mathbb{E}_{\mathcal{M}\sim p(\mathcal{M})}[\mathbb{E}_{s^t,\mathbf{u}^t}[\sum_t \gamma^t r(s^t,\mathbf{u}^t)]]$.

\subsection{Configuration of the Task}
In this section, we develop a simplified 2D dynamical model for multiple quadrupedal robots pulling a cable-towed load.
These robots are connected to the load through $n$ cables, each with length $\{l_i\}_{i\in \mathcal{A}}$. The load is a square with an edge length of $l_l$ and a mass of $m_l$. There are $m$ obstacles in the environment, all of which are square with an edge length $l_o$. Specifically, at time step $t$, the system has the following configuration space: 

\begin{eqnarray}
\begin{aligned}
\mathbf{q^t_l} &:= [x^t_l, y^t_l, \theta^t_l]^T \in SE(2),  \\
\mathbf{q^t_{r_i}} &:= [x^t_{r_i}, y^t_{r_i}, \theta^t_{r_i}]^T \in SE(2), i=1,..,n, \\
\mathbf{q_{o_j}} &:= [x_{o_j}, y_{o_j}, \theta_{o_j}]^T \in SE(2), j=1,..,m, \\
\end{aligned}
\end{eqnarray}
where $\mathbf{q^t_l}$ represents the load's configuration, $\mathbf{q^t_{r_i}}$ represents the $i$-th robot's configuration, and $\mathbf{q_{o_j}}$ represents the $j$-th obstacle's configuration in the world frame at time step $t$. The $i$-th cable is connected between the $i$-th attachment point on the load surface and the centroid of the $i$-th robot. The coordinates of the $i$-th attachment point at time step $t$ correspond to the center of one of the edges of the load and can be denoted as 
$[x^t_l+\frac{l_l}{2}\cos(\theta^t_l+\theta_{a_i}), y^t_l+\frac{l_l}{2}\sin(\theta^t_l+\theta_{a_i})]^T \in \mathbb{R}^2$, where $\theta_{a_i}$ takes values in the set $\{0, \frac{\pi}{2}, \pi, \frac{3\pi}{2}\}$. Each edge of the load can support multiple attachment points, allowing for the attachment of more than four cables. We further define the coordinates of the destination as $\mathbf{q_d}= [x_d,y_d]^T \in \mathbb{R}^2$. 
Partially observable observations at time step $t$ is defined as $\mathbf{o}_i^t = [{\mathbf{q}_l^t}^T, {\mathbf{q}_{r_i}^t}^T, {\mathbf{r}^t_{i,\text{neighbor}}}^T, {\mathbf{o}^t_{i,\text{neighbor}}}^T, \mathbf{q}_d^T]^T$, where $\mathbf{r}^t_{i,\text{neighbor}}$ and $\mathbf{o}^t_{i,\text{neighbor}}$ represent all other robots or obstacles that are within an $L1$ distance less than $d_{\text{thresh}}$ from robot $i$, where $d_{\text{thresh}}$ represents the maximum visible distance.

\section{Hierarchical System for Navigation of Multiple Quadrupedal Robots}

In this section, we develop a hierarchical robotic system designed to solve the multi-robot navigation task. We outline its structure, which includes a global planner for the load, as well as decentralized planners and locomotion controllers for the robots.

\begin{figure}[t]
\begin{center}
\subfloat{
\begin{centering}
\includegraphics[width=0.7\linewidth]{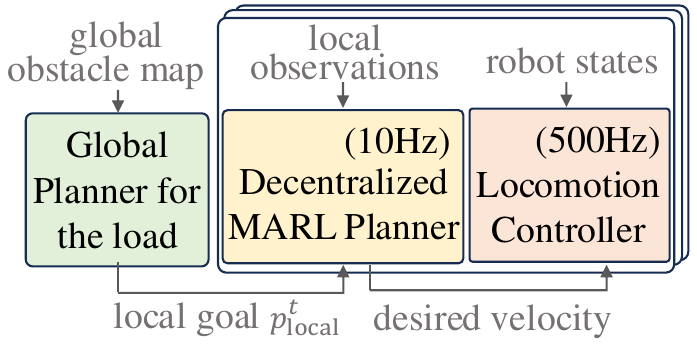}
\end{centering}
}
\end{center}
\vspace{-3mm}
\caption{The proposed hierarchical robotic system consists of three main components: the global planner for the load, decentralized MARL planners, and locomotion controllers. The global planner handles long-horizon planning, generating a collision-free trajectory for the load in cluttered environments. The decentralized MARL planners operate independently on each robot, managing local multi-robot collaboration by generating velocity commands based on local observations and local goals. The locomotion controller, running at a higher frequency, computes joint torques for the quadruped robots using MPC, ensuring robust locomotion in response to environmental changes and terrain variations.}
\label{fig:main_method} 
\end{figure}

\subsubsection{Overall Architecture}
The proposed hierarchical robotic system, depicted in Fig.~\ref{fig:main_method}, comprises three components: a global planner for the load, a decentralized MARL-based planner, and a locomotion controller. The global planner focuses on long-horizon planning, guiding the multi-robot system through cluttered environments while avoiding obstacles. The decentralized planner handles multi-robot collaboration and complex interactions, running independently on each robot. The locomotion controller ensures robust locomotion for quadruped robots, enabling them to navigate uneven terrain and withstand environmental perturbations. It also runs independently on each robot.

\subsubsection{Global Planner for the Load}
The key insight is that while Multi-Agent Path finding (MAPF) performs well in long-horizon tasks, it faces challenges in scaling to large-scale multi-agent collaboration~\cite{sartoretti2019primal} due to the exponential growth of the search space when accounting for collision avoidance between agents. In contrast, MARL algorithms are effective at handling collaboration but face challenges with long-horizon tasks~\cite{liu2024imagine}. 
To address this, we use A$^*$ to plan the trajectory of the load. A$^*$ is executed at the start of each episode, taking the global obstacle map, the load's position, and the goal position as inputs. It outputs a collision-free path from the load to the goal. Note that the robots' positions are not considered during this step. 
Next, a set-point 1.8 meters ahead of the load's current position along the A$^*$ path is selected as the local goal $p_{\text{local}}^t$ for the decentralized planner. As the system approaches the destination, the target point \( q_d \) serves as the local goal.

\subsubsection{Decentralized Planner}
The decentralized planners run independently on each robot, handling inter-robot collaboration. They run at a frequency of 10 Hz. They take local observations and the local goal $p_{\text{local}}^t$ provided by the global planner as inputs and generate high-level velocity commands for the robots. These decentralized planners generate desired velocities for the robots in three degrees of freedom: linear velocities in the $x$ and $y$ directions and a yaw angular velocity. The decentralized planners are trained using a MARL algorithm, which is detailed in Section~\ref{sec:MARL_algo}.

\subsubsection{Locomotion Controller}
The desired velocity commands output by the decentralized planners are sent to the locomotion controller, which uses MPC to compute the joint torques for the quadruped robots to track the desired velocities. The locomotion controller operates at a much higher frequency of 500 Hz to ensure that it can quickly adjust the robot's behavior in response to environmental perturbations or uneven terrain.

\section{MARL Algorithm for the Decentralized Planner}
\label{sec:MARL_algo}

In this section, we propose an algorithm for training a decentralized planner capable of controlling any number of robots under the CTDE framework. In Section~\ref{sec:CTDE}, we analyze the challenges of applying CTDE to environments with varying team-size, providing a high-level overview of the algorithm's design. Following this, we present a detailed explanation of the algorithm along with its implementation details.

\subsection{CTDE for Scenarios with a Varying Number of Robots}
\label{sec:CTDE}

In our work, we leverage the CTDE framework to train MARL agents. The key idea behind CTDE is to use a critic network that has access to comprehensive global information, such as other agents' states or privileged states. This allows the critic to better model the environment and teammates, improving training stability, and enhancing overall performance. However, the critic is used only during training and is discarded at execution time. The planner only takes local observations as input. This ensures that during execution time, the system operates in a decentralized manner, with each agent relying solely on its local observation for decision-making. This makes the framework suitable for domains with a large number of robots.

While CTDE has shown effectiveness in fixed-team-size scenarios, applying it to environments with a varying number of robots introduces the following challenges.

\subsubsection{Adapting to Different Environment Configuration}

When the environment configuration varies, such as changes in the number of obstacles or robots, the length of the local observations also changes. Adapting to these varying configurations requires adjustments to both decentralized execution and centralized training.
For decentralized execution, we carefully design the observation space to ensure that each robot’s partial observation has the same format and dimension, regardless of team size. This design allows the same decentralized actor network to control teams of any size during execution, eliminating the need for re-training or adaptation at runtime.
However, the same approach cannot be directly applied during centralized training, since the carefully designed input only represents partial local observations.
For centralized training, the global state input to the critic network, which includes all robots' local observations and privileged information, varies in dimension as the number of robots changes. To address this, we train separate critic networks for each team size. This implies that larger teams require critic networks with greater capacity to handle the increased system complexity. The privileged information is only used during training.

\subsubsection{Managing Increasing Multi-Robot System Complexity}

The complexity of multi-robot systems grows exponentially with the number of robots. Each team size requires a distinct critic network, making simultaneous training for varying team sizes impractical. To address this, we propose a multi-stage training pipeline. In each stage, training is conducted on a fixed team size scenario. It begins with a single-robot scenario and gradually increases the number of robots. This curriculum-based approach accelerates training by incrementally adding complexity. We trained the policy using teams of one to four robots and deployed the resulting policy, which is capable of controlling teams of any size within this range, for evaluation.

\subsubsection{Mitigating Catastrophic Forgetting and Loss of Plasticity}

The multi-stage training pipeline, while effective, introduces challenges associated with catastrophic forgetting and loss of plasticity, common issues in continual learning frameworks.
Catastrophic forgetting~\cite{french1999catastrophic} occurs when policies trained on new scenarios overwrite the knowledge from previous tasks. For instance, after training on a four-robot scenario, the policy may forget how to control two-robot teams. To prevent this, we introduce a multi-agent knowledge distillation loss during multi-stage training. This loss ensures that the updated policies do not deviate significantly from previously learned ones, preserving past knowledge.
Loss of plasticity~\cite{dohare2024loss} refers to the diminished ability to learn new tasks after extensive training, often due to convergence to local optima. This issue limits scalability to larger team sizes. To maintain agents' ability to learn new knowledge, we periodically reset the critic network during training, drawing inspiration from prior work~\cite{nikishin2022primacy,ma2023revisiting}. These resets help the system retain its flexibility and learn effectively in increasingly complex scenarios.

\begin{figure}[t]
\centerline{\includegraphics[width=0.85\linewidth]{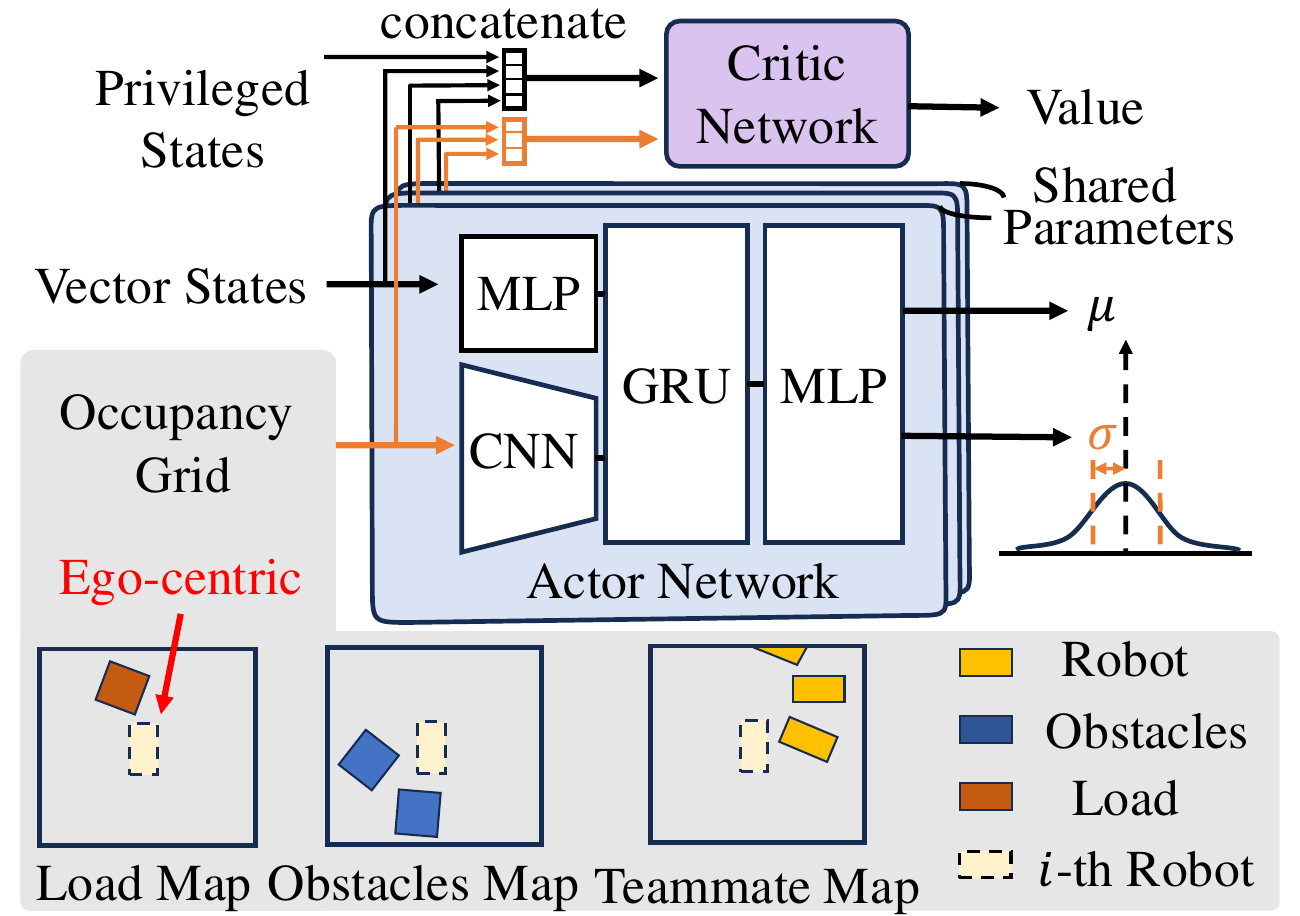}}
\caption{Design of the MARL-based decentralized planner and an illustration of the $i$-th robot's local occupancy grid map. 
The actor network processes multi-modal inputs to generate desired velocities for each robot. 
Inputs include vector states and ego-centric local occupancy grid maps, with a key design feature being their dimension-invariance to accommodate variable team sizes.
The critic network processes the global state, formed by concatenating all local observations and privileged information such as values of domain randomization variables. While sharing the actor's architecture, the critic employs larger hidden layers to handle the complexity of the global state.
The bottom figure illustrates the $i$-th robot’s ego-centric local occupancy grid maps, which include a load map indicating the load’s position, an obstacle map, showing the nearby obstacles, and a teammate map reflecting the positions of other robots.
The $i$-th robot is always positioned at the center of the map and oriented upward.}
\label{fig:RL_framework}
\end{figure}

\subsection{Design of the Decentralized Planner}

This section introduces the decentralized planner designed to manage a multi-robot system. We provide a comprehensive overview of the design and architecture of the actor and critic networks, which serve as the core components of the proposed multi-stage MARL framework.

\subsubsection{Actor Network}

Our decentralized planner, also referred to as the actor network $\pi_\theta$, is modeled as a deep neural network parameterized by $\theta$. The actor network architecture is designed to be consistent in size regardless of the team size, enabling it to control a variable number of robots effectively. In multi-robot scenarios, each robot operates its own instance of the actor network, with all networks sharing the same parameters. Despite sharing parameters, the robots exhibit distinct behaviors based on their unique local observations.

As illustrated in Fig.~\ref{fig:RL_framework}, the input to the $i$-th robot's policy at time step $t$ comprises two components: a vector state and local occupancy grid maps. The vector state includes the $i$-th robot's position, yaw (expressed as sine and cosine values to ensure continuity), the position of the load, and the coordinates of the attachment point of the $i$-th cable on the load, which corresponds to the center of one of the load's surfaces.
The local occupancy grid maps capture nearby robots, obstacles, and the load, with each feature encoded in a separate map. These ego-centric maps are centered on the $i$-th robot, which is positioned at the center and oriented upward. The maps have a fixed size of $3.42\,\text{m} \times 3.42\,\text{m}$.

At each time step $t$, the output of the $i$-th robot's actor network is a pair of vectors, $\boldsymbol{\mu} \in \mathbb{R}^3$ and $\boldsymbol{\sigma} \in \mathbb{R}^3$, representing the mean and variance of three Gaussian distributions. In general, a well-trained policy exhibits low variance. To avoid numerical instability, the variance is clipped at a predefined lower bound. During execution, an action is randomly sampled from these Gaussian distributions, with the resulting values corresponding to the desired sagittal and lateral velocities in the $i$-th robot's local frame and the angular velocity for turning. 
For a well-trained model, sampling from the Gaussian distributions and selecting the mean value of these distributions have a similar effect as the variance is low.

\subsubsection{Critic Network}

The critic network takes global states as input.
We describe the construction of the global state as follows. The global state is the concatenation of all local observations. The local occupancy grid maps are concatenated at the channel level. The vector state is also concatenated with privileged states, including the values of the domain randomization variables and the current time-step.

\subsubsection{Model Architecture}

As shown in Fig.~\ref{fig:RL_framework}, the $\pi_\theta$ architecture is designed to process multi-modal inputs. A multilayer perceptron (MLP) handles the vector state, while a 2D convolutional neural network (CNN) processes the local occupancy grid maps. The base MLP consists of two hidden layers, each with 128 ReLU units. 
The 2D CNN encoder is composed of four hidden layers, configured as \([ \text{kernel size}, \text{filter size}, \text{stride size} ]\): \([5, 8, 3]\), \([3, 16, 2]\), \([3, 32, 2]\), and \([3, 64, 2]\). Each layer uses ReLU activation without padding. The features extracted from the MLP and CNN are concatenated to form a unified feature vector, which is then passed as input to a recurrent network based on a Gated Recurrent Unit (GRU).
The GRU can implicitly capture I/O history, enabling it to model complex temporal dependencies and patterns. 
It consists of a single recurrent layer with orthogonal weight initialization and layer normalization, ensuring robust training and stability.
The final MLP consists of two separate heads: one for predicting the mean and the other for the variance of the Gaussian distribution. Each head contains two hidden layers, each with 128 ReLU units.
The critic network shares the same architectural design as the actor network but uses hidden layers that are twice as large. This increased capacity allows the critic to effectively process the more complex global state information.

\begin{figure*}[t]
\begin{center}
\subfloat{
\begin{centering}
\includegraphics[width=0.65\linewidth]{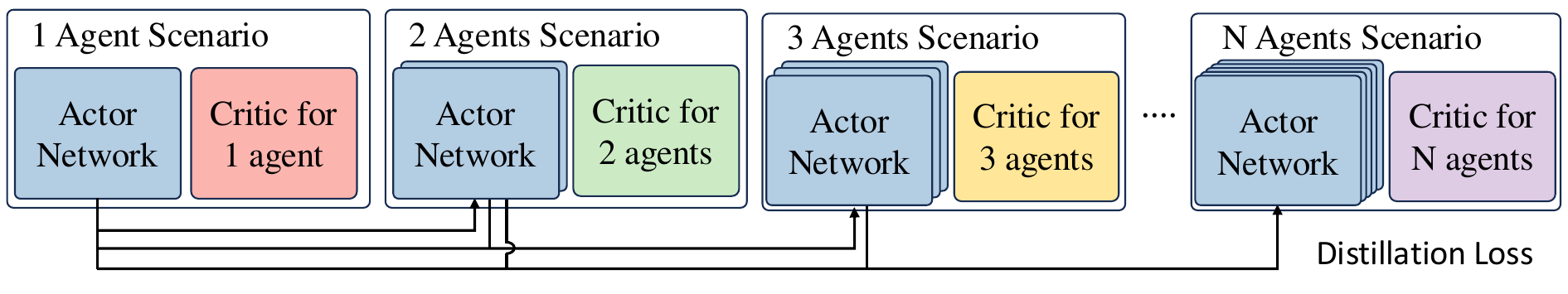}
\end{centering}
}
\end{center}
\vspace{-3mm}
\caption{The multi-stage training pipeline is designed to train the decentralized planner across varying team sizes. Starting with a single-robot scenario, the number of robots is gradually increased at each stage. The policy is trained using MAPPO, with the actor network's parameters initialized from the previous stage. This approach allows the network to generalize across varying team sizes. To prevent catastrophic forgetting, a multi-agent knowledge distillation loss is added to the actor’s loss function. The critic network is periodically reset and trained from scratch at the start of each stage.}
\label{fig:training} 
\end{figure*}

\subsection{Multi-Stage Training Framework}

Building on the introduction of the decentralized planner, we now present a general framework for training it with multi-agent reinforcement learning. 

\subsubsection{Overall Training Pipeline}

We propose a multi-stage training pipeline, as illustrated in Fig.~\ref{fig:training}, where each stage focuses on a specific team size.
The training begins with a single-robot scenario and gradually introduces more robots into the environment.
That is, in the first stage, \( p(\mathcal{M}) = [1,0,\dots,0] \); in the second stage, \( p(\mathcal{M}) = [0,1,\dots,0] \); and so on.
At each stage, the policy is trained using MAPPO~\cite{yu2022surprising}. The actor network's parameters are initialized using those from the previous stage, with random initialization only applied in the first stage. Since the observation length is designed to be independent of the number of robots, the same actor network can be consistently reused across stages with different team sizes.
To address catastrophic forgetting, we incorporate a multi-agent knowledge distillation loss into the actor's loss function:
\begin{equation}
L = L_{\text{MAPPO}} + \frac{\beta}{T-1}\sum_{k=1}^{T-1}D_{KL}\big(p_k(\cdot)||p_T(\cdot)\big),
\end{equation}
where \( L_{\text{MAPPO}} \) denotes the MAPPO actor loss, $\pi_k$ is the policy trained on scenario with $k$ robots, \( p_k(\cdot) \) represents the probability distribution of $\pi_k$, \( p_T(\cdot) \) represents the probability distribution of the current policy $\pi_T$, and \( \beta \) is a hyper-parameter that controls the weight of the regularization term.
The critic network is reset periodically, which means that the critic network is randomly initialized and trained from scratch at the beginning of each stage.
The reward functions and the termination condition are shared among different stages, as outlined below.

\subsubsection{Reward and Episode Design}

We now define the reward function and the task's termination condition. These are designed to encourage the multi-robot system to navigate the load to a designated goal position as quickly as possible while avoiding collisions. The reward function and termination condition are shared among all robots.
The reward at time-step t, denoted as $r_t$, consists of two main components: (1) task completion, and (2) local goal tracking. 
The task completion reward is a binary value. A positive reward is given to the team if the load’s position is sufficiently close to the goal:
\begin{equation}
r_{ex} = \begin{cases}
1, ||\mathbf{q_d}-[x^t_l,y^t_l]^T||_2<T_d \\
0, \text{otherwise},
\end{cases}
\end{equation}
where $T_d$ is a hyperparameter set to 0.5 m for this task.
However, this binary reward is sparse, which can hinder exploration. To address this, we introduced an additional intrinsic reward for local goal tracking to guide exploration.
The intrinsic reward is designed to reflect the load’s progress toward the local goal. Specifically, it is computed as the distance the load moves towards the local goal within a single time step. The intrinsic reward at time step $t$ is given by:
\begin{equation}
r_{in} = ||p^t_{\text{local}}-[x^t_l,y^t_l]^T||_2 - ||p^t_{\text{local}}-[x^{t+1}_l,y^{t+1}_l]^T||_2.
\end{equation}
The final reward function is given by $r = \alpha r_{\text{ex}} + r_{\text{in}}$, where $\alpha$ is a weighting factor set to the robot’s maximum velocity.

Two conditions terminate the episode: (1) timeout and (2) collision. 
At the start of each episode, the distance between the load and the goal, $d$, is calculated. The maximum time-step $T_{\text{max}}$ is proportional to $d$, defined as $T_{\text{max}} = 8 d_t + 20$ seconds. The constant $20$ is added to allow robots to reorient themselves if they are not initially facing the goal.
The second termination condition is any collision involving any of the robots (with other robots, obstacles, the load) or between the load and obstacles.
Unlike previous work, our reward function does not include a penalty term for collision avoidance, as this is implicitly addressed through the termination condition.

\begin{figure}[t]
\centerline{\includegraphics[width=0.99\linewidth]{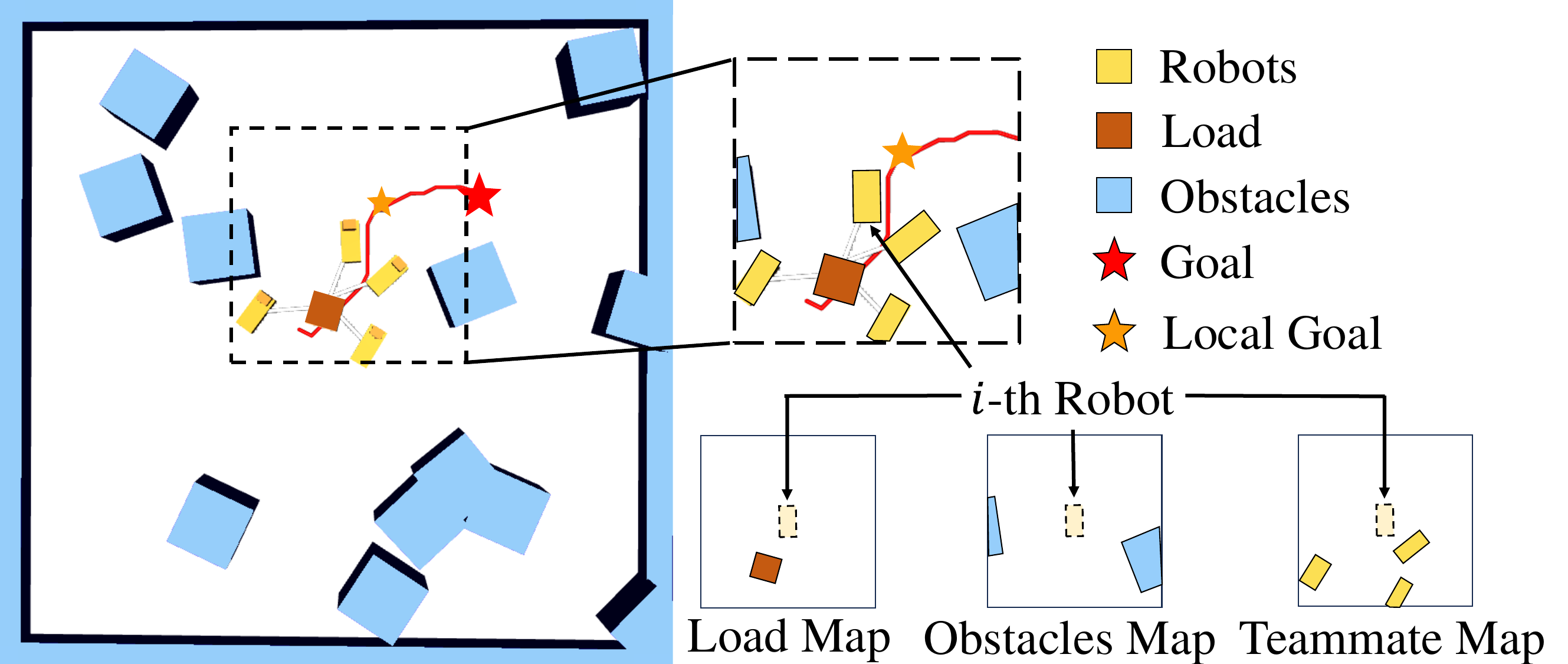}}
\caption{A screenshot of the MuJoCo training environment depicts blue blocks as obstacles and walls, yellow blocks as robots, a brown box as the load, and red lines indicating the global planning path for the load. A red star marks the goal, while an orange star represents the local goal. The figure on the right displays the local occupancy grid map of the $i$-th robot, consisting of three ego-centric maps: the load map showing the load's position, the obstacles map reflecting nearby obstacles, and the teammate map indicating the positions of nearby robots. The $i$-th robot itself is excluded from these local occupancy maps.}
\label{fig:mujoco}
\end{figure}

\subsection{Simplified Environment Dynamics for RL Training}

As illustrated in Fig.~\ref{fig:mujoco}, we utilized MuJoCo as our training environment. The map was a \(10\,\text{m} \times 10\,\text{m}\) square, enclosed by walls along its boundaries. Inside the arena, there were ten \(1\,\text{m} \times 1\,\text{m}\) obstacles, which were allowed to overlap. Cables connecting the robots and the load were modeled using tendons implemented in MuJoCo.

In the simulation, legged robots were simplified as box-shaped entities controlled by PD controllers with three degrees of freedom: motion along the x-axis, motion along the y-axis, and angular rotation. This simplification was necessary because simulating detailed legged robot dynamics is computationally expensive, and RL requires large amounts of data. To address the sim-to-real gap introduced by this simplification, we applied domain randomization to the PD gain. Empirical results demonstrate that this approach effectively bridges the sim-to-real gap and enables the use of different types of robots within the team. For example, in our experiments, we successfully used both A1 and Go1 robots together.

\subsection{Domain Randomization}

We incorporate domain randomization parameters into the simulation environment to train a policy that is robust and generalizable to uncertainties in measurements, environment dynamics, and configurations. This domain randomization approach facilitates the successful transfer of policies from simulation to real-world scenarios, where such uncertainties are inherent.

To address uncertainties in environment dynamics modeling, we randomize various parameters, including the PD gains of the robot, friction coefficients, load weight, and cable length. These parameters are randomly set at the start of each episode and remain fixed throughout that episode. Additionally, these dynamic modeling parameters are included as part of the input to the critic network to aid in effective learning.

To manage measurement uncertainty, we introduce simulated noise to observable states, including the positions of the load, robots, and obstacles. This simulates real-world sensor inaccuracies and helps the policy adapt to noisy inputs.

Finally, given that the proposed MARL system is designed for a navigation task in cluttered terrain, it is crucial to simulate diverse environment configurations. At the start of each episode, the load, robots, obstacles, and the goal's position and orientation are randomly initialized. To ensure the task is feasible, we use the A\(^*\) algorithm to verify that a valid, unblocked path exists from the load to the goal.

\subsection{Training Details}

For each scenario with different numbers of robots, we generated 1,024 randomized configurations as described in the previous section. The experiments were conducted on a machine equipped with 128 GB of RAM, a 128-core CPU, and a GeForce RTX 3090 GPU. 
The model was trained to control one to four robots, with a total of 50 million time steps. Specifically, 15 million time steps were allocated to scenarios involving one and four robots, while 10 million time steps were used for two- and three-robot scenarios. The entire training process took eight days to complete.

\section{Simulation Validation}

In this section, we begin by outlining the setup of our validation simulation environment. Subsequently, we present the simulation results to address the following research questions (RQ): 
\textbf{RQ1:} How effectively does the proposed algorithm perform in multi-robot collaboration tasks?
\textbf{RQ2:} Does the multi-robot system forget previously learned skills during continual learning, and does the proposed multi-agent knowledge distillation loss mitigate this issue?
\textbf{RQ3:} How critical is the choice of centralized training?
\textbf{RQ4:} How efficient is the decentralized planner compared to the centralized one?

\subsection{Simulated Environment Setup}

We evaluated our multi-robot system in the simulated environment to verify the proposed algorithm in multi-robot collaboration tasks. The testing scenarios, which were not seen during training, were randomly initialized.
All other settings are identical to those in the training environment.
Our evaluation included 1024 distinct maps for each number of robot scenarios, with both the mean and standard deviation of results reported. 

\begin{table}[t]
\caption{Evaluation of the proposed algorithm on multi-robot collaboration tasks\label{table:main_results}}
\begin{center}
\begin{tabular}{|c|cccc|}
\hline
Number Agents & \textbf{1}& \textbf{2}& \textbf{3} & \textbf{4} \\
\hline
Avg. Speed (m/s) & 0.27$\pm$0.18 & 0.26$\pm$0.17 & 0.24$\pm$0.15 & 0.24$\pm$0.14 \\
Cosine Similarity & 1.00$\pm$0.00 & 0.52$\pm$0.36 & 0.39$\pm$0.17 & 0.28$\pm$0.12 \\
\hline
Success rate (\%) & 80.98 & 71.46 & 71.90 & 68.94 \\
\hline
\end{tabular}
\end{center}
\end{table}

\begin{figure}[t]
\begin{center}
\subfloat{
\begin{centering}
\includegraphics[width=0.95\linewidth]{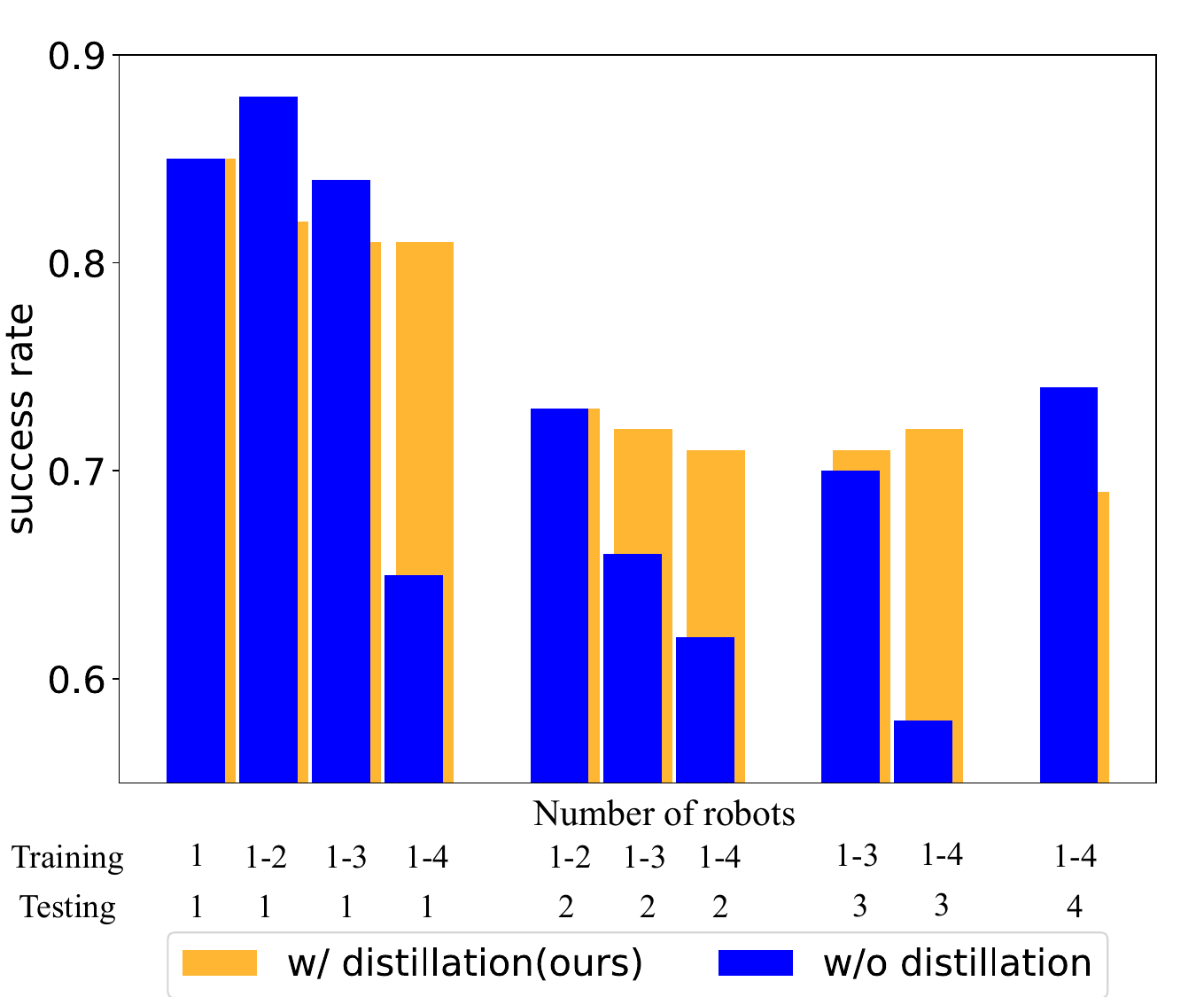}
\end{centering}
}
\end{center}
\vspace{-3mm}
\caption{The effect of multi-agent knowledge distillation loss in mitigating catastrophic forgetting is evaluated in multi-robot scenarios. Performance is compared between setups with and without distillation loss, showing consistent skill retention when the distillation loss is applied.
The ``Training" row indicates the scenarios in which the model has been trained, such as: ``1-3" means the model has been trained in scenarios with a single robot, two robots, and three robots. The ``Testing" row represents the number of robot scenarios in which the model is being tested.
For instance, the first blue (dark) bar represents the success rate of an agent trained and tested in the single-robot scenario, demonstrating high performance. Conversely, the fourth blue bar illustrates the success rate of an agent trained on one-to-four-robot scenarios but tested in the single-robot scenario, where performance drops significantly due to catastrophic forgetting. In contrast, the yellow (light) bars, representing our method, consistently show high success rates across all scenarios. This result underscores the effectiveness of knowledge distillation in preserving skills and preventing forgetting.}
\label{fig:knowledge_distillation} 
\end{figure}

\subsection{Multi-Robot Collaboration (RQ1)}

As shown in the success rate row in TABLE~\ref{table:main_results}, the proposed method effectively controls with one to four robots to navigate a cable-towed load through cluttered spaces, using a single model across all four scenarios. 
The success rate in a single-robot scenario is approximately 80\%. This is because, in some randomly generated maps, robots may start by facing loads, obstacles, or be positioned close to other robots, increasing the likelihood of collisions. This issue can be mitigated in real-world experiments by adding buffers around obstacles, robots, and the load.
The success rate is approximately 70\% in scenarios involving two to four robots, highlighting the inherently greater difficulty of multi-agent collaboration.
The average speed (Avg. Speed) reflects the algorithm’s efficiency. 
Empirical results indicate that the average speed remains consistent across scenarios with varying numbers of robots. This indicates that the team of robots collaborates effectively, as a team of non-cooperative robots would become stuck, hindering each other's movements.
The cosine similarity indicates the alignment of the robots’ heading directions. Higher cosine similarity values suggest successful collaboration among the robots. In contrast, values lower than zero indicate opposing forces, causing the system to get stuck. Notably, aiming for extremely high cosine similarity is impractical, as it could lead to robot collisions, especially as the number of robots increases.

\begin{figure}[t]
\begin{center}
\subfloat{
\begin{centering}
\includegraphics[width=0.95\linewidth]{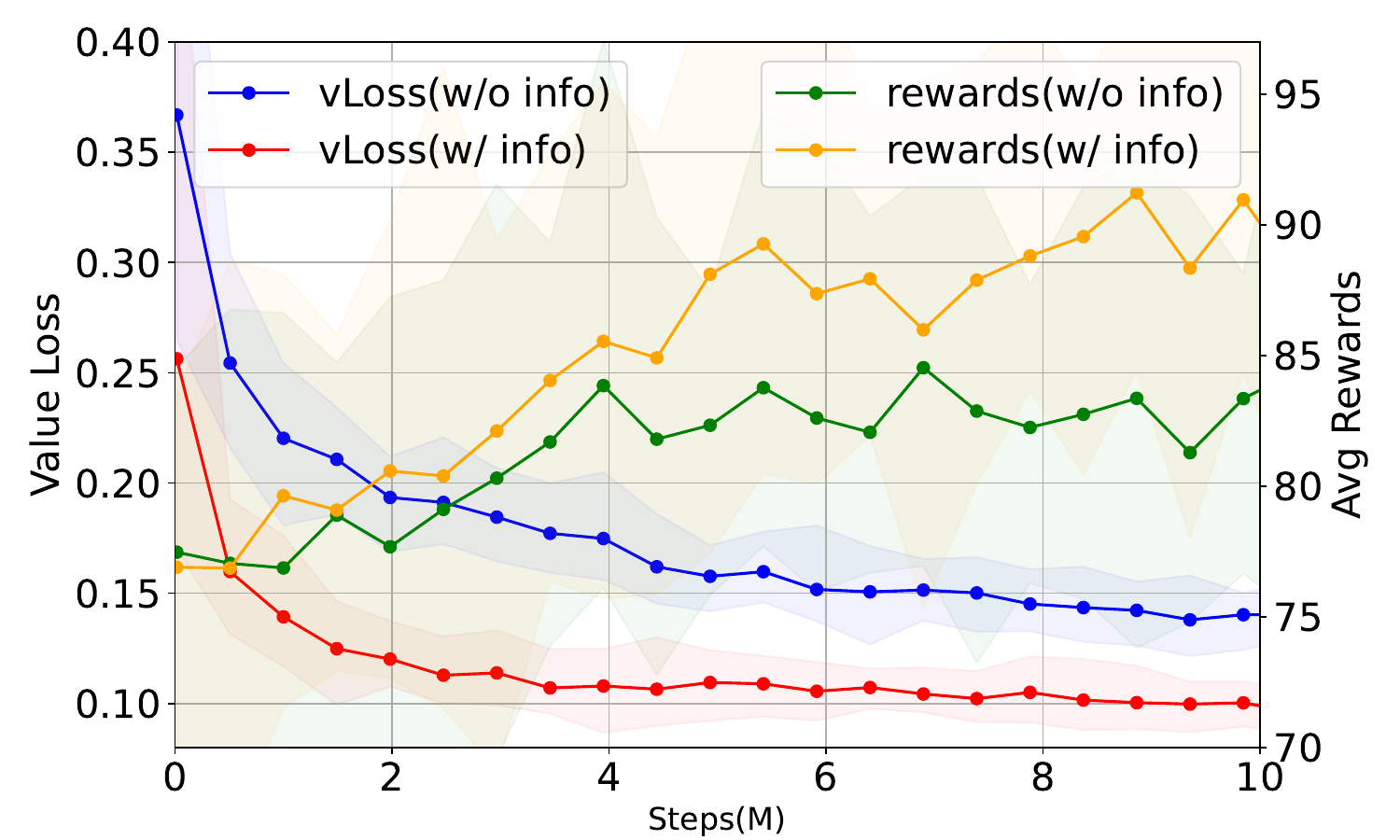}
\end{centering}
}
\end{center}
\vspace{-3mm}
\caption{Performance comparison of the critic network with and without access to global and privileged state information in a two-agent multi-robot collaboration scenario. Comparing the blue and red curves indicates that training the critic with global information (red) reduces value loss. Similarly, comparing the yellow and green curves shows that incorporating global information (yellow) leads to higher average rewards.}
\label{fig:ctde} 
\end{figure}

During training, the proposed framework heavily relies on the local goals generated by the global planner for the load, as these goals are used both as a reward and as part of the local observation. However, during testing, the multi-robot system may choose paths that deviate from the trajectory suggested by the global planner. This discrepancy arises because the global planner considers only obstacles and the load when generating trajectories. In some cases, the suggested trajectory may lead to potential collisions between robots. To avoid such conflicts, the multi-robot system adapts by selecting alternative paths. This demonstrates the adaptability of the proposed MARL-based decentralized planner and alleviates concerns about the hierarchical robotics system's dependence on the path planner for the load.
\textbf{Take away message:} The proposed MARL algorithm can effectively control teams of one to four robots to complete the collaborative navigation task. Additionally, the algorithm is not heavily reliant on the output generated by the path planner for the load.

\subsection{Multi-agent Knowledge Distillation (RQ2)}

Empirical results indicate that continual training of our multi-robot system on scenarios involving increasing numbers of robots can lead to a phenomenon known as "catastrophic forgetting," where previously learned knowledge is lost. As shown in Fig.~\ref{fig:knowledge_distillation}, the performance of the blue (dark) bar (representing training without knowledge distillation) declines as the system is trained on more challenging scenarios.
For example, the first four bars on the left are tested in the one-robot scenario, while the rightmost bar in this group is trained on scenarios involving one to four robots, and the leftmost bar is trained only on the one-robot scenario. Results show a 20\% performance drop in the rightmost bar compared to the leftmost bar (both blue), demonstrating the forgetting effect when knowledge distillation is absent.
In contrast, the proposed multi-agent knowledge distillation loss effectively mitigates this issue. As illustrated in Fig.~\ref{fig:knowledge_distillation}, incorporating multi-agent knowledge distillation loss enhances performance when testing and training scenarios differ. The first bar in each test environment group shows that performance remains consistent, with or without the distillation loss, indicating that distillation does not negatively impact training results in the latest iteration. The remaining bars illustrate that training with the distillation loss helps agents retain skills associated with controlling fewer robots, which were acquired in earlier training stages. Additionally, we emphasize the importance of our observation space design, which ensures that the local observation dimension remains invariant to the number of robots, enabling continuous learning.
\textbf{Take away message:} Multi-agent knowledge distillation loss prevents the policy from forgetting previously learned skills, making it an essential design component for enabling a single decentralized planner to control varying numbers of robots effectively.

\subsection{Centralized Training (RQ3)} 
We conducted an ablation study on centralized training by training the critic with and without global and privileged state information. As shown in Fig.~\ref{fig:ctde}, the model trained with access to global and privileged state information exhibits lower value loss and higher average episode rewards. These results support our claim that privileged information improves training stability. Specifically, the value function loss is significantly lower than that of independent value functions, demonstrating that the proposed multi-robot collaboration task cannot be solved as multiple independent single-agent tasks. The evaluation was conducted in a two-agent scenario. Note that the critic network parameters are randomly initialized at the start of each iteration, making its accuracy and sample efficiency crucial in this setting.
\textbf{Takeaway Message:} Training the critic network with privileged state and global information improves its accuracy and performance, addressing the limitations of independent learning in solving this multi-robot collaboration task.


\begin{figure}[t]
\begin{center}
\subfloat{
\begin{centering}
\includegraphics[width=0.95\linewidth]{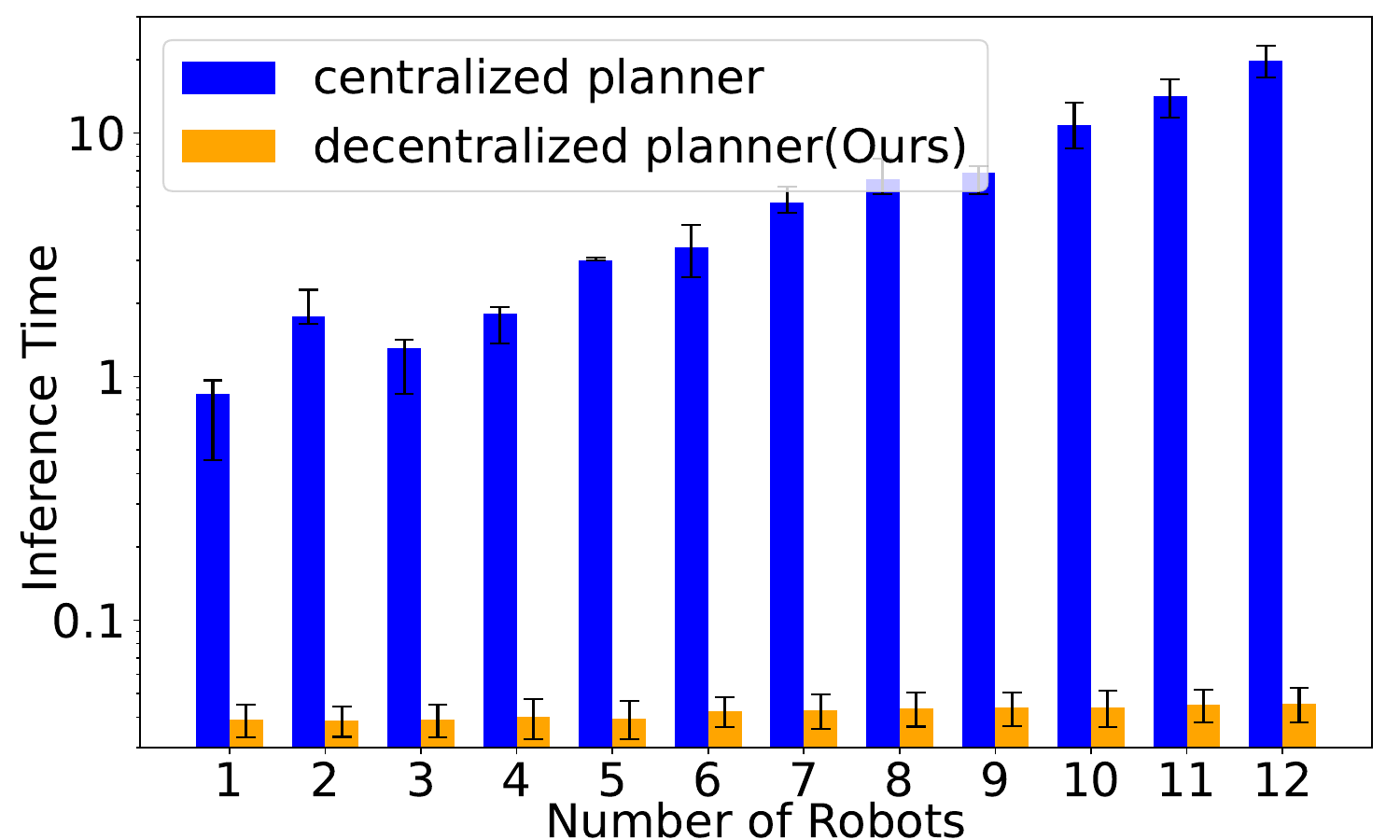}
\end{centering}
}
\end{center}
\vspace{-3mm}
\caption{Comparison of inference time between our decentralized planner and the centralized planning approach from~\cite{9830869}. While our method maintains a constant inference time as the number of robots increases, the centralized approach exhibits exponential growth. This highlights the efficiency of the decentralized planner. The y-axis is shown on a logarithmic scale to emphasize the performance differences.}
\label{fig:infer_time} 
\end{figure}

\subsection{Decentralized Execution (RQ4)}

We assessed the inference time required by our multi-robot system. Fig.~\ref{fig:infer_time} compares the inference time of our method with that of \cite{9830869} , which utilizes a centralized planning method with some heuristic parallelism. Results show that our method's inference time remains constant with the number of robots, while the centralized method scales exponentially. Although \cite{9830869} speeds up inference by simplifying and parallelizing the multi-robot system, it still faces exponential growth in shared action space, resulting in over 20 seconds of planning time in 12 robots scenario—unsuitable for real-time application. In contrast, our method requires less than 0.1 seconds of planning time for 12 robots. 
\textbf{Takeaway Message:} A decentralized planner makes decisions based solely on local observations. As a result, the inference time remains constant regardless of the number of robots, enabling real-time planning even for large team sizes.

\subsection{Scalability (RQ4)}
\begin{figure}[t]
\begin{center}
\subfloat{
\begin{centering}
\includegraphics[width=0.95\linewidth]{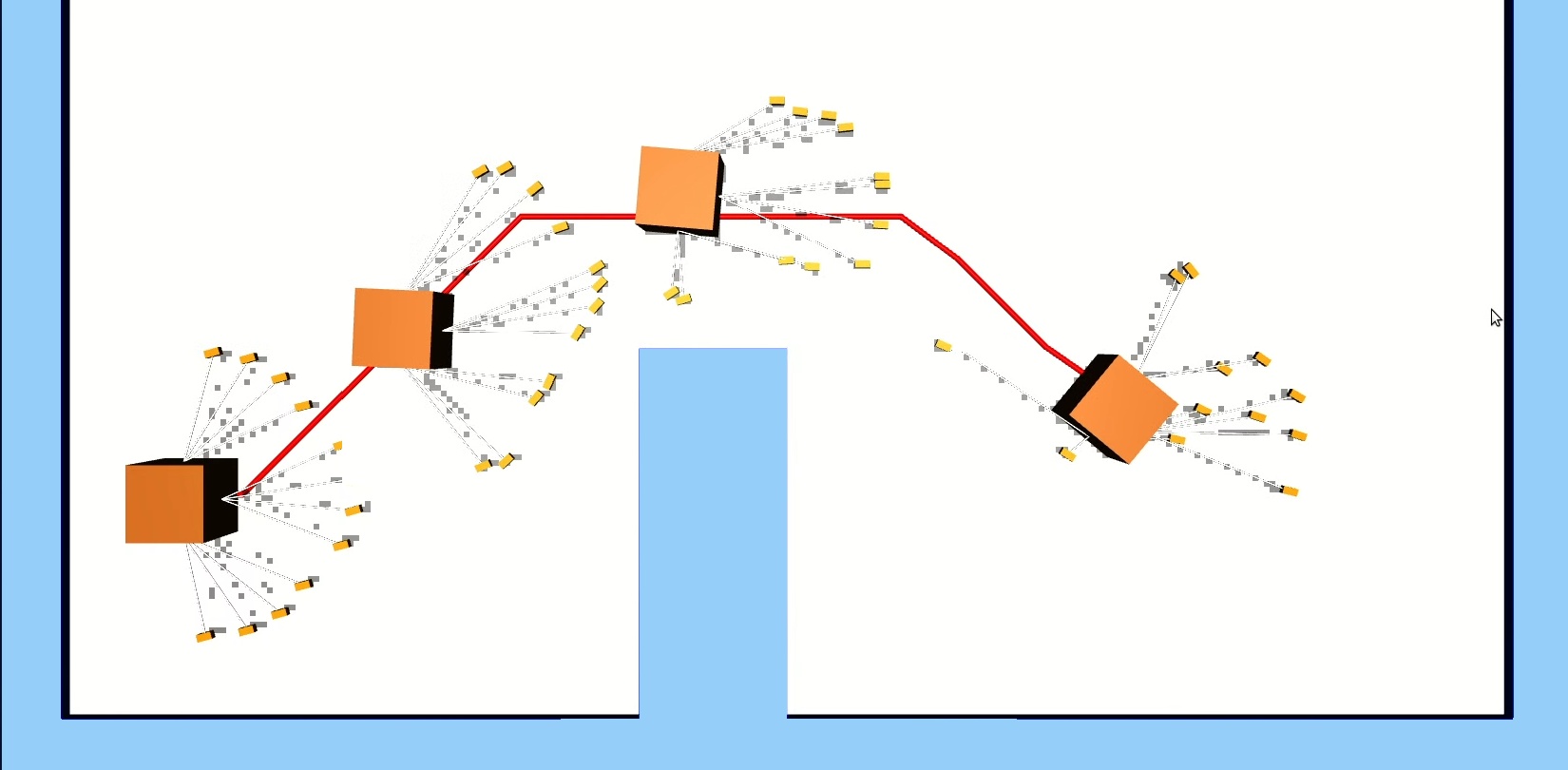}
\end{centering}
}
\end{center}
\vspace{-3mm}
\caption{Illustration of the multi-robot navigation task in a simulated environment. Each robot, shown as a yellow rectangle, is responsible for towing a cable-connected load (brown box) to its assigned target, the red curve's rightmost point, and performing a challenging U-turn maneuver while avoiding blue obstacles. This experiment demonstrates the scalability of the proposed method. The \href{https://youtu.be/GkGldcfQi9k}{video} showcases how our method enables a single model to control teams of varying sizes, from one to twelve robots. The figure above illustrates the trajectory generated by our method in a 12-robot scenario.}
\label{fig:twelve_robots} 
\end{figure}

We evaluated our approach on a larger multi-robot system to assess its scalability and effectiveness in complex scenarios. Inspired by the setup in the previous work~\cite{9830869}, we implemented and tested our method within a simulated environment with up to 12 robots. The training pipeline described in Section~\ref{sec:MARL_algo} was scaled accordingly to accommodate up to 12 agents. 
As demonstrated in the \href{https://youtu.be/GkGldcfQi9k}{video}, our method successfully uses a single model to control teams of varying sizes, from one to twelve robots, enabling them to complete a challenging U-turn scenario.
Figure~\ref{fig:twelve_robots} illustrates how our approach controls twelve robots to coordinate and execute the U-turn maneuver. The current version of this experiment is configured such that both training and testing phases are conducted within the same environment.
\textbf{Takeaway Message:} This experiment highlights the scalability of our approach, demonstrating its capability to effectively control and coordinate a large team of 12 robots in a simulated environment.

\section{Experiments}

In this section, we first describe the experimental setup and then evaluate the proposed multi-robot system across a range of challenging real-world scenarios. 
The system can effectively control the robots in the real world without requiring additional tuning after being trained in the simulation environment. 
The experiments focus on assessing two key aspects of the proposed method: (1) The flexibility of the multi-robot system to handle varying team sizes. (2) The robustness of the system to environmental perturbations and uncertainties in the real-world. The results are recorded in the accompanying \href{https://youtu.be/GkGldcfQi9k}{video}.

\subsection{Experimental Setup}

We used four quadruped robots to conduct real world experiments: one Unitree A1 and three Go1s. The room used for experiments was approximately $4 \times 6$ meters while the load and obstacles were all $0.5 \times 0.5 \times 0.5 \ m^3$ boxes. The robots were attached to the load with 1 meter cables. We varied the weight by adding the $1.2$ kg Go1 batteries and $0.87$ kg A1 batteries to the box. The robots, load, and obstacles were localized using a Vicon motion capture system. The same decentralized planner, trained on scenarios of one to four robots, is used for testing across all scenarios.

\subsection{\textit{L-turn} Experiments}

\begin{figure*}[t]
\begin{center}
\subfloat{
\begin{centering}
\includegraphics[width=0.98\linewidth]{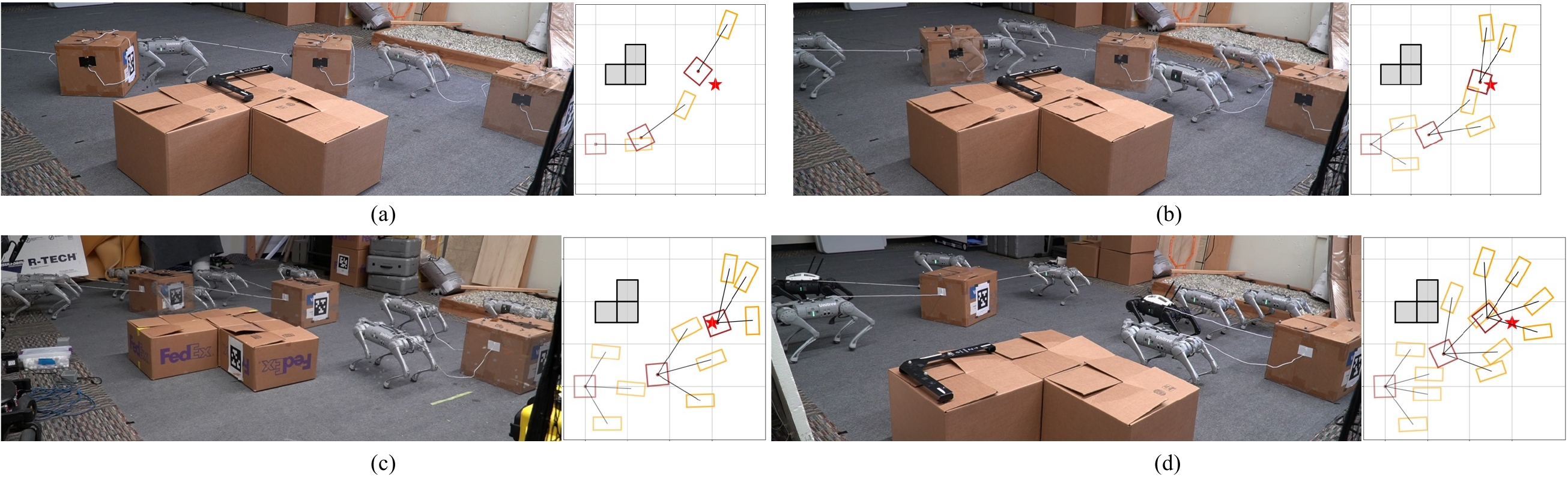}
\end{centering}
}
\end{center}
\vspace{-3mm}
\caption{The performance of our multi-robot system in a \textit{L-Turn} scenario, where robots must move forward and execute a 90-degree left turn while avoiding obstacles. The task is completed with one to four robots using the same decentralized planner. The system does not know the number of robots and relies on local observations to infer collaborative strategy. In the four-robot scenario, experiments were conducted with one Unitree A1 and three Go1s. Results demonstrate that domain randomization enables the system to generalize across robots of different types and dynamics. (a)-(d) present the \textit{L-Turn} scenario involving one to four robots. Right: Top-down view of the map. Left: Stacked frames showing the beginning, middle, and end positions of the task, with later positions displayed in lower transparency.}
\label{fig:iturn}
\end{figure*}

We first evaluated our multi-robot system in a \textit{L-Turn} scenario, where the goal is to move forward and turn left while avoiding obstacles.
The robots initially move forward before executing a 90-degree turn, adding complexity as the inner and outer robots must coordinate at different speeds. 
As shown in Fig.~\ref{fig:iturn}, our algorithm successfully controls one to four robots to complete the task using the same decentralized planner. 

During testing, the system is not informed of the number of robots, yet the robots demonstrate the ability to infer collaborative behavior based on their local observations. 
For instance, in the single-robot \textit{L-Turn} scenario, the robot begins turning left at \(x=2\). In contrast, in the three-robot scenario, the middle robot starts turning left at \(x=3\) to allow more space for the inner robots to navigate the turn. 
The ego-centric local occupancy grid map spans \(3.42\text{m} \times 3.42\text{m}\), ensuring that, while not all robots are visible at all times, each robot can observe nearby teammates and make decisions to collaborate effectively. 

In the four-robot scenario, we conducted experiments using one Unitree A1 and three Go1. The results show that the proposed domain randomization enables the multi-robot system to generalize effectively, controlling robots of different types despite variations in size and dynamics.

\subsection{\textit{Narrow-Gap} Experiments}

To evaluate the impact of incorporating cables, we tested our multi-robot system in a \textit{Narrow-Gap} scenario. 
In this scenario, the robots face a row of obstacles with a narrow 2.3 m gap. Due to terrain constraints, the system cannot pass through the gap with all cables taut, requiring the robots to adopt a contracted formation. After passing the obstacles, the system transitions back to its normal formation.

When the robots pull the cable-towed load, the lateral forces from the load may cause collisions between the robots. Experimental results show that the low-level locomotion controller effectively handles these perturbations, and the decentralized planner adjusts the robots' directions to avoid collisions. Operating the decentralized planner at 10 Hz is crucial, as lower frequencies often lead to collisions. This highlights the importance of decentralized execution, as the inference time of a centralized planner increases exponentially with the number of robots, making it impractical for real-time control of larger teams.

\begin{figure*}[t]
\begin{center}
\subfloat{
\begin{centering}
\includegraphics[width=0.98\linewidth]{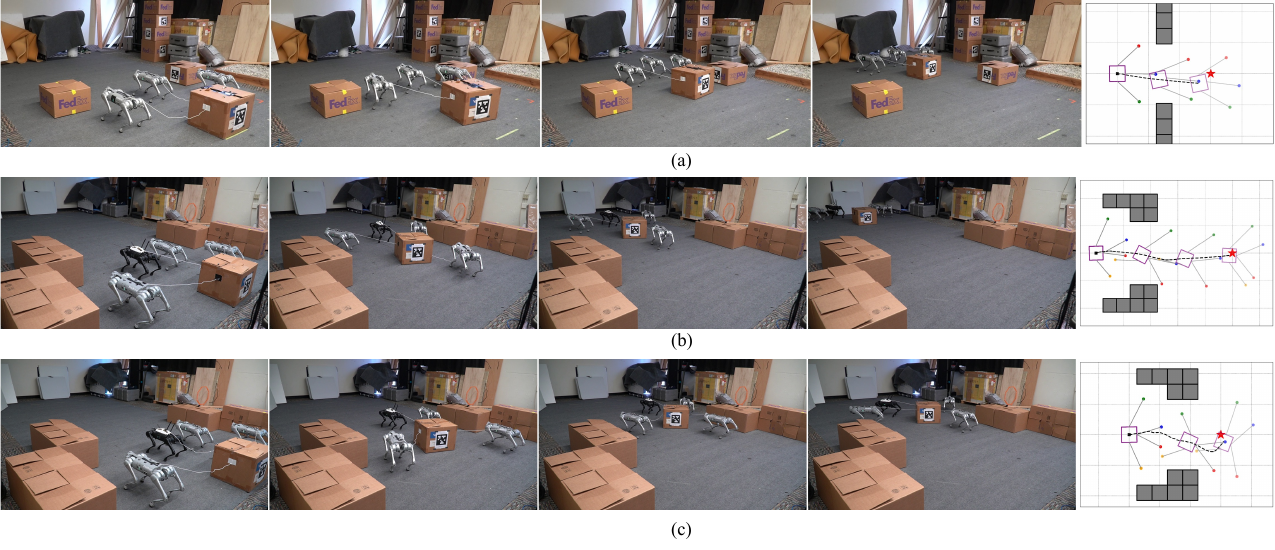}
\end{centering}
}
\end{center}
\vspace{-3mm}
\caption{The performance of our multi-robot system in the \textit{Narrow-Gap} scenario. (a) Three-robot navigation. (b) Four-robot navigation with a 4.6 kg load. (c) Four-robot navigation with a lighter load (1 kg). In \textit{Narrow-Gap} scenario, robots navigate a narrow 2.3 m gap between obstacles while pulling a cable-towed load. To pass through the gap, the system adopts a contracted formation, adjusting the cables between taut and slack states. (b)(c) showed the different formations of the multi-robot system when navigating loads of varying weights. For lighter loads, two robots lead the team, while two follow. For heavier loads, three robots take the lead, with one following.}
\label{fig:windows}
\end{figure*}

We tested the method with three and four robots using the same decentralized planner. As shown in Fig.~\ref{fig:windows}, our system successfully navigates the narrow gap by dynamically adjusting the cables between taut and slack states. The results indicate that our decentralized planner requires 33.03 seconds to collaboratively control three robots to navigate the load over a distance of 3 meters, whereas previous methods relying on a centralized planner~\cite{9830869}, took approximately 49.52 seconds (1.5 times longer) to achieve the same task.

\subsection{Adaptivity to Unknown Load Weights}

During testing, the decentralized planner is not provided with the load's weight, so we expect the multi-robot system to automatically adjust its formation to different load weights. To verify the adaptivity to unknown load weights of the proposed method, we tested it under two load conditions: 1 kg and 4.6 kg, using the four-robot \textit{Narrow-Gap} scenario, where all the configurations are unseen during testing time. As shown in Fig.~\ref{fig:windows} (c), with a 1 kg load, a single robot was sufficient to pull the load. In this case, two robots pulled from the front while the other two followed, forming a configuration that is well-suited for navigating narrow spaces and reducing the risk of collision with obstacles. As shown in Fig.~\ref{fig:windows} (b), when the load weight was increased to 4.6 kg, at least three robots were needed to pull it. The multi-robot system adjusted their formation so that three robots pulled through the narrow gap while one followed behind. 

\begin{figure*}[t]
\begin{center}
\subfloat{
\begin{centering}
\includegraphics[width=0.95\linewidth]{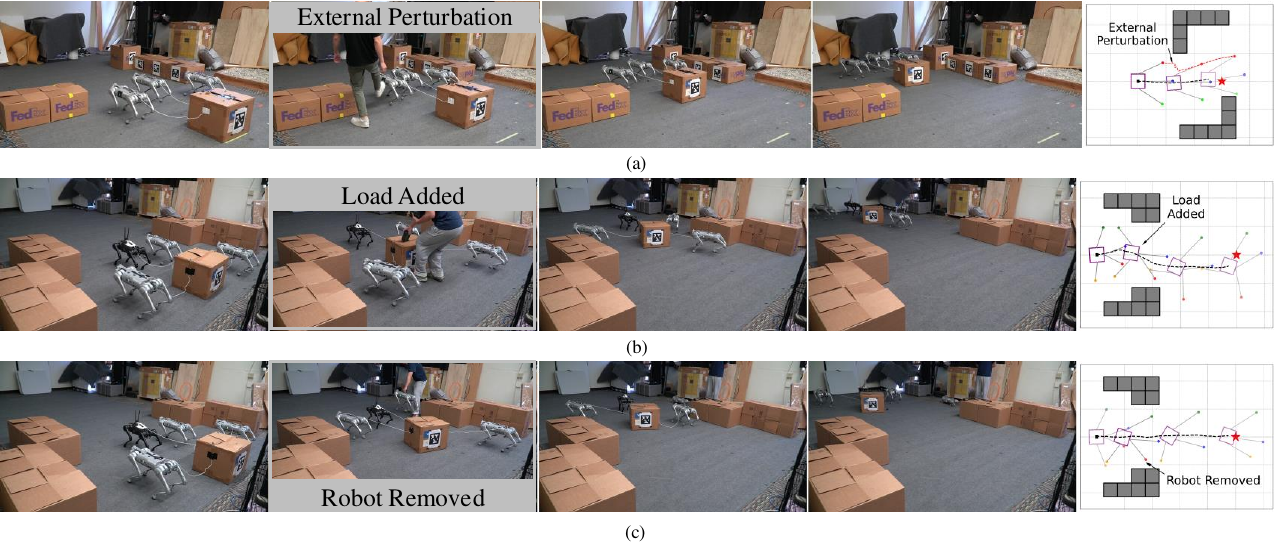}
\end{centering}
}
\end{center}
\vspace{-3mm}
\caption{Adaptation of the Multi-Robot System to Dynamic Environmental and Configuration Change
(a) \textit{External Perturbation} scenario: During testing, a robot is perturbed and displaced, but it quickly corrects its trajectory without colliding with others.
(b) \textit{Changing Weights} scenario: An additional 3.8 kg load is added during testing. The system dynamically adjusts its formation in real time to accommodate the increased weight.
(c) \textit{Changing Number of Robots} scenario: One robot is removed during testing. The system seamlessly transitions from four robots to three, dynamically adjusting its formation to efficiently navigate the load and avoid obstacles.}
\label{fig:adaptation}
\end{figure*}

We further tested the multi-robot system in a more advanced scenario, where we manually added an additional 3.6 kg to the load after the episode had started. This increase in weight required three robots to pull the load, as opposed to the two robots that were sufficient before the load was added. As shown in Fig.~\ref{fig:adaptation}, the multi-robot system can adapt its formation in real time: before the load is added, two robots lead the team, with two following; after the load is added, three robots lead and one follows. The formation transformation occurs on-the-fly, and the multi-robot system successfully avoids collisions between robots and obstacles during this adaptation.

These results demonstrate that, despite the absence of any explicit collaboration term in the reward function or prior knowledge about collaboration strategies, the multi-robot system is able to automatically learn the most effective collaboration pattern to adapt to different environmental configurations, such as different load weights.

\subsection{Robustness to Environment Perturbation}

The experiment discussed in the previous paragraph is promising, though not surprising, as load weight variation was included in the domain randomization objective. This means that during training, the multi-robot system was explicitly taught to handle different load weights.

In this section, we demonstrate the multi-robot system's remarkable ability to quickly adapt to changes in the environment. Specifically, we tested the system in the \textit{External Perturbation} scenario. In this setup, one of the robots was manually perturbed and displaced by 0.36 meters within 1.5 seconds during the middle of the test. The results show that the disturbed robot quickly corrected its trajectory to avoid collisions with other robots.

The robustness of the multi-robot system to environment perturbation, despite this type of perturbation not being seen during training, can be attributed to three factors. First, the decentralized planner is trained using a MARL algorithm, which is generalizable to unseen environments. Second, the decentralized planner enables the real-time control of the multi-robot system, allowing quick recovery from disturbances. Lastly, the MPC-based locomotion controller ensures robustness to environmental perturbations.

\subsection{Adaptivity to Change of Team-Size}

To verify the multi-robot system’s ability to collaborate with varying numbers of robots, we tested it in a more advanced \textit{Changing Number of Robots} scenario. In this scenario, we began testing with a system of four robots. Halfway through testing, one robot was removed.  
As illustrated in Fig.~\ref{fig:adaptation} (c), the multi-robot system effectively adapted its formation. It transitioned from a formation where three robots were leading and one was following, to a new formation where all remaining robots led. 
Note that the system was not provided with any direct information about the number of robots in the team. Instead, the decentralized planner adjusted its behavior based solely on the changes in the local observations. For instance, when a robot was removed, it disappeared from the local occupancy grid map, allowing the remaining robots to update their collaboration strategy accordingly. This result highlights the effectiveness of training a model that can seamlessly adapt to changes in team size, improving the system's flexibility in handling dynamic robot configurations.

\section{Conclusion}

This paper introduces a decentralized MARL framework for collaborative navigation of cable-towed loads by multiple quadrupedal robots. Our hierarchical robotic system combines global path planning with a decentralized MARL planner, enabling robots to make decisions based solely on local observations. Experiments in simulation and real-world scenarios validate the system’s scalability, adaptability to varying robot count, and robustness to changes in load weight and environmental perturbation, demonstrating its effectiveness for complex multi-robot collaboration tasks.

While the proposed system demonstrates promising results, there are still limitations to address: 
(1) Dependency on a Global Obstacle Map: The global planner for the load currently relies on a pre-defined global obstacle map, which may not be feasible in environments with partial observability. To overcome this, future work will involve running the A$^*$ algorithm periodically and constructing a global obstacle map incrementally based on local observations. This approach aims to enhance the system's applicability in real-world scenarios with limited prior information.
(2) Decoupling of Locomotion Controller and Decentralized Planner: The hierarchical system currently decouples the low-level locomotion controller from the mid-level decentralized planner. Although this separation simplifies training and implementation, it may limit the system's overall performance. Future research will explore end-to-end training methods that jointly optimize both components, potentially improving coordination and robustness in multi-robot systems.
(3) The goal of this paper is to leverage a large number of small, general-purpose quadrupeds rather than rely on a single, larger, special-purpose robot to solve real-world tasks. Consequently, we assume that all robots in the system are homogeneous, meaning they share the same roles and capabilities. This assumption does not apply to scenarios where robots have heterogeneous roles—for instance, some robots might be designated for navigation while others focus on exploring the environment~\cite{9349130}.



\section*{Acknowledgments}
This work was supported in part by Design of Robustly Implementable Autonomous and Intelligent Machines, Defense Advanced Research Projects Agency award number HR00112490425, and in part by The Robotics and AI Institute. The authors would like to thank Prof. Claire Tomlin for giving access to the motion capture system, as well as Lizhi Yang and Qiayuan Liao for their help with the experiments. The authors also thank Chenyu Yang, Dr. Akshara Rai, and Dr. Jun Zeng for insightful discussions. K. Sreenath has financial interests in The Robotics and AI Institute. He and the company may benefit from the commercialization of the results of this research.



 

\bibliographystyle{IEEEtran}
\bibliography{IEEEabrv,ref}  

\end{document}